\documentclass[10pt,twocolumn,letterpaper]{article}

\usepackage{cvpr}
\usepackage{times}
\usepackage{epsfig}
\usepackage{graphicx}
\usepackage{amsmath}
\usepackage{amssymb}
\usepackage{textcomp}
\usepackage{multirow}
\usepackage{siunitx}
\usepackage{booktabs}
\usepackage{cite}
\usepackage{xcolor}
\usepackage{caption}

\newcommand{\commentMR}[1]{\textcolor{black}{#1}}
\newcommand{\commentPO}[1]{\textcolor{black}{#1}}

\usepackage{authblk}

\usepackage[breaklinks=true,bookmarks=false]{hyperref}
\usepackage{cleveref}

\cvprfinalcopy 

\def\cvprPaperID{7} 

\newcommand*\phantomrel[1]{\mathrel{\phantom{#1}}}
\DeclareMathOperator*{\argmax}{arg\,max}

\ifcvprfinal\fi
\begin{document}

\title{Detection and Retrieval of Out-of-Distribution Objects in Semantic Segmentation}

\author[1]{Philipp Oberdiek}
\author[2]{Matthias Rottmann}
\author[1]{Gernot A. Fink}
\affil[1]{Department of Computer Science, TU Dortmund University}
\affil[2]{School of Mathematics and Natural Sciences, University of Wuppertal}

\renewcommand\Authands{ and }


\twocolumn[{%
\renewcommand\twocolumn[1][]{#1}%
\maketitle
\begin{center}
    \centering
    \includegraphics[width=\textwidth,height=9cm,keepaspectratio]{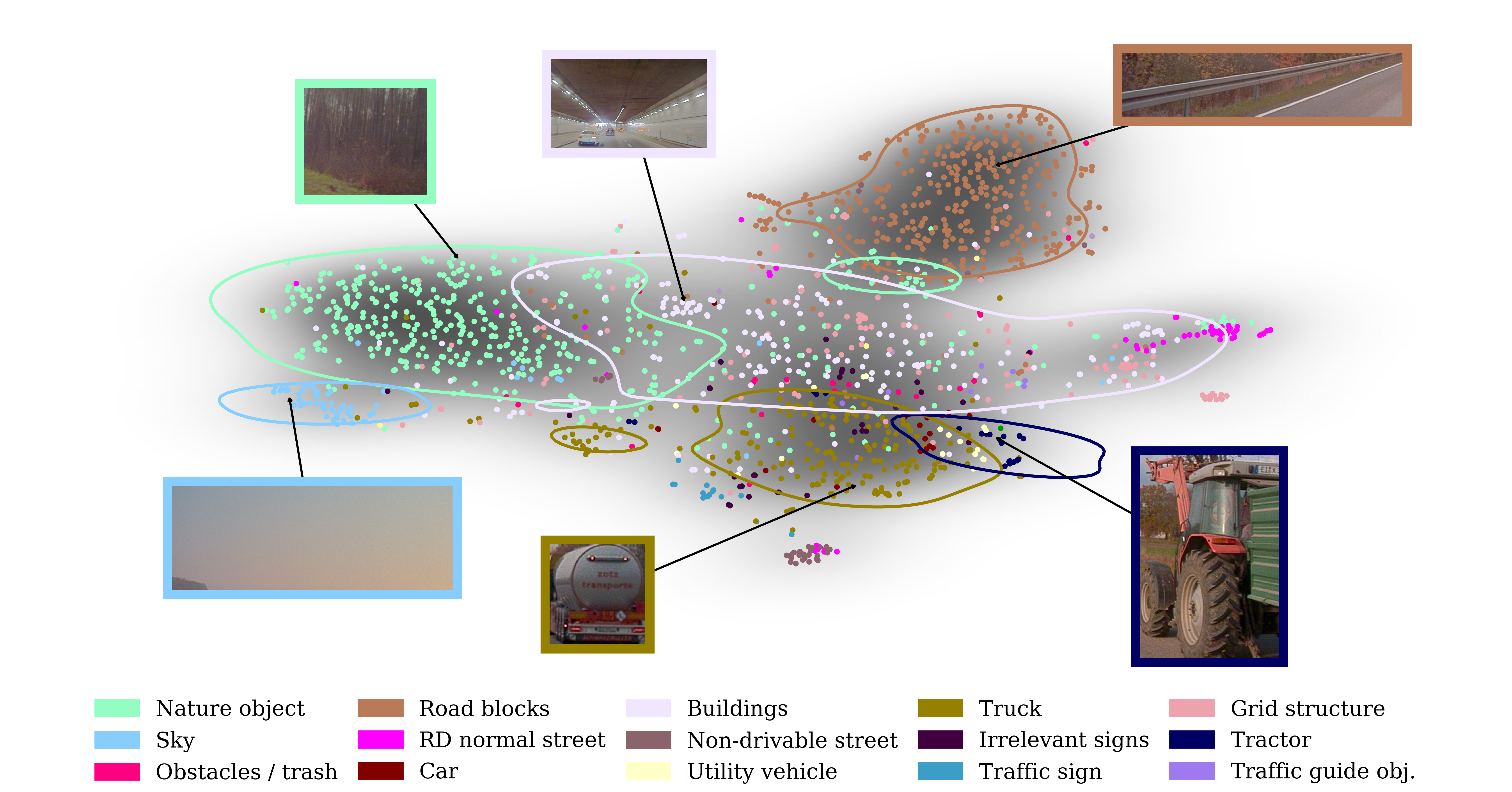}
    \captionof{figure}{Embedding space of unknown and badly segmented objects based on ResNet152 features. Darker regions symbolize higher global density. Contour lines are \textit{regions of highest density} \cite{hdr} of gaussian kernel density estimates of the data conditioned to the classes depicted on the thumbnails. Bandwidth selection for the gaussian kernel density estimates is done using \textit{Scott's rule} \cite{scottsrule} and we select $\alpha = 0.2$ for the \textit{highest density regions}. Dimensionality reduction has been performed using PCA down to 50 dimensions followed by \textit{t-SNE} \cite{tsne} with perplexity of $30$, early exaggeration of $12$ and learning rate of $200$.}
    \label{fig:embedding_space_density}
\end{center}%
}]

\thispagestyle{empty}

\begin{abstract}
   When deploying deep learning technology in self-driving cars, deep neural networks are constantly exposed to domain shifts. These include, \eg, changes in weather conditions, time of day, and long-term temporal shift. In this work we utilize a deep neural network trained on the Cityscapes dataset containing urban street scenes and infer images from a different dataset, the A2D2 dataset, containing also countryside and highway images. We present a novel pipeline for semantic segmenation that detects out-of-distribution (OOD) segments by means of the deep neural network's prediction and performs image retrieval after feature extraction and dimensionality reduction on image patches. In our experiments we demonstrate that the deployed OOD approach is suitable for detecting out-of-distribution concepts. Furthermore, we evaluate the image patch retrieval qualitatively as well as quantitatively by means of the semi-compatible A2D2 ground truth and obtain mAP values of up to $52.2\%$.
\end{abstract}

\section{Introduction}

\commentMR{The advances of convolutional neural networks (CNNs) in the recent years enabled the use of machine learning for complex computer vision tasks that had been considered out of reach before. Among them is the semantic segmentation that facilitates complex scene understanding \cite{LongSD15}.}
Applications like \eg autonomous driving, medical imaging or surveillance are problem domains that induce a high risk and lead to fatal consequences \commentMR{when using CNNs in an autonomous and unsupervised fashion.}
Thus it is of utmost importance to monitor \commentMR{CNNs} and ask for human intervention when \commentMR{questionable} predictions are detected \cite{HendrycksG17}.

In general, there are many aspects of a \commentMR{machine learning pipeline for computer vision} that \commentMR{require} supervision, not necessarily by humans. Starting with data collection at the very beginning it is of high importance to collect a sample of the visual world that represents the sub environment \commentMR{for which a CNN's deployment is desired}.
As the visual world has basically an infinite variability, \commentMR{a sufficient representation (in particular in safety relevant scenarios) is not easy to accomplish}. Additionally, the acquisition of annotation can be an expensive and time consuming task. A variety of publications presented different possible approaches. While some publications try to reduce the cost of label acquisition with techniques like \emph{semi supervised learning} \cite{PapandreouCMY15}, \emph{weakly supervised learning} \cite{BearmanRFL16, PapandreouCMY15, LinDJHS16} or \emph{active learning} \cite{RottmannKG18, SenerS18}, more recent works on \emph{self-supervised learning} \cite{DoerschGE15, KolesnikovZB19} try to extract visual features without requiring labeled data. 
\commentMR{Furthermore, generating synthetic data for training neural networks is also considered.} Publications in this direction include the rendering of highly realistic images \cite{synscapes} or using generative adversarial networks (GANs) for generation and augmentation of already collected data \cite{Sankaranarayanan18}.

\commentMR{Similar to the data acquisition phase, monitoring is also required when a trained model is deployed in the (open) real world -- which is under constant change over time and space}. As we cannot assume that \commentMR{available} training samples represent the \commentMR{target real world environment well, it is highly relevant to} track possible prediction failures and situations that are completely new to the model at hand. \commentMR{Important research areas that tackle these problems include \emph{uncertainty} or \emph{confidence} estimation as well as} \emph{out-of-distribution} (OOD) detection. Works in the field of \emph{uncertainty} or \emph{confidence} estimation include Bayesian methods \cite{KendallBC17,ZhaoSZQ18,GalG16}, ensemble methods \cite{Lakshminarayanan17} as well as approaches that acquire information from intermediate layers of the network or from its predictions to train a second model that \commentMR{serves} as confidence estimator \cite{metaseg,OberdiekRG18,HendrycksG17,DeVries18}. The task of OOD detection has been broadly studied \commentMR{for} image recognition \cite{HendrycksG17,LiangLS17,OberdiekRG18,DeVries18} and most of these methods are applicable to the problem of semantic image segmentation. 
Approaches specifically designed for semantic segmentation include e.g.~\cite{BevandicKOS19,Mehrtash19}. \commentPO{Both try to measure confidence on pixel level which is in contrast to the method used in this work. The authors of \cite{BevandicKOS19} utilize shared convolutional features to predict segmentation confidence with an additional output branch. Using a negative dataset as a proxy for OOD objects, they train their auxiliary model to predict model confidence. In \cite{Mehrtash19} the authors propose to use ensembles of models to calibrate the prediction confidence. This is however computationally expensive, especially for state-of-the-art semantic segmentation models in the context of street scene segmentation.}

During autonomous driving, even a simple change of location can result in a severe domain shift resulting in unseen objects. \commentMR{Additionally, the real world is subject to continuous transformation. Therefore} it is indispensable to update deep learning models regularly.
This results in a continuous feedback loop between the previously described steps of a machine learning pipeline, the data acquisition stage and the deployment phase. Our work makes important contributions to a more efficient workflow of this process. Using a \commentMR{meta classification and regression approach termed \emph{MetaSeg}} \cite{metaseg} to find unknown objects, we can group the detected entities into visually and semantically related groups in order to enhance data exploration in the \commentMR{presence} of domain shift. Using predicted segmentation masks and image retrieval within newly collected data, our approach can be used to find classes that may be underrepresented or missing in the training dataset. This knowledge can be used for example to improve the existing model by partly labeling novel object classes and including them into the next training round. In summary the contributions of this work are as follows:
\begin{enumerate}
    \item We show that MetaSeg predicts the intersection over union of out of domain samples reliably.
    \item \commentMR{Using MetaSeg we demonstrate that we are able to detect unknown object classes.}
    \item By extracting visual features we are able to group the found entities into an embedding space with semantically related neighborhoods.
    \item We perform an evaluation on the task of image retrieval with a variety of common deep learning architectures as feature extractors.
\end{enumerate}
\commentMR{To the best of our knowledge, this is the first work that reliably detects OOD samples in semantic segmentation and reveals their semantic similarity.}

\commentMR{The remainder of this work is structured as follows:} 
\commentMR{In \cref{sec:oodd} and \cref{sec:retrieval}} we describe the theoretical foundations of our OOD detection and retrieval pipeline. \commentMR{Using the Cityscapes dataset \cite{cityscapes} as source domain and the A2D2 \cite{a2d2Dataset} dataset as target domain (out of domain sample) we demonstrate} in \cref{subsec:metaseg_eval} that we are able to reliably detect unknown objects in the presence of domain shift. Complementing the meta segmentation analysis we conduct experiments on an image retrieval task and present in \cref{subsec:retrieval_eval} qualitative and quantitative results to demonstrate that this technique can be used to enhance data exploration for semantic image segmentation.


\section{\commentMR{Out-of-Distribution Detection}} \label{sec:oodd}
\commentMR{Under the premise that objects of unknown classes mostly cause suspicious predictions, we can quantify this effect and use it for OOD detection.
Therefore we deploy an approach that estimates segmentation quality for each predicted segment by means of statistical properties. This approach is termed \emph{MetaSeg} \cite{metaseg} and was developed further in \cite{RottmannS19,Maag2019}. Based on a structured dataset of metrics that aggregate dispersion measures of the softmax output as well as geometric properties of each predicted segment, a regression model is trained to predict the segmentation quality (in terms of segment-wise intersection over union (IoU) with the ground truth, also known as the Jaccard index \cite{Jaccard12similarityCoefficient}). To this end, we train a small fully-connected neural network on the set of metrics solely corresponding to in-distribution data.}
\commentMR{In what follows we describe this procedure in more detail, starting with the construction of metrics as proposed in \cite{RottmannS19}:}

Given the output $f_z(y|x,w)$ of the semantic segmentation model for input $x$, weights $w$ and pixel $z$ over class labels $y\in\mathcal{C}=\{y_1,\ldots,y_K\}$, we \commentMR{compute the pixel-wise classification} \emph{entropy}
\begin{equation}
    E_z(x,w) = -\frac{1}{\log (K)}\sum_{y\in\mathcal{C}}f_z(y|x,w)\log f_z(y|x,w)\label{eq:entropy} \, ,
\end{equation}
the \emph{probability margin}
\begin{align}
    \begin{split}
        M_z(x,w) & = 1 - f_z(\hat{y}_z(x,w)|x,w) \label{eq:probdiff}\\
        & \phantomrel{=} + \max_{y\in\mathcal{C}\setminus\{\hat{y}_z(x,w)|x,w)\}} f_z(y|x,w) \, ,
    \end{split}
\end{align}
and the \emph{variation ratio} 
\begin{equation}
    V_z(x, w) = 1 - f_z(\hat{y}_z(x,w)|x,w) \label{eq:varrat} \, ,
\end{equation}
with $\hat{y}_z(x,w)=\argmax_{y\in\mathcal{C}} f_z(y|x,w)$ being the predicted class of pixel $z$.

After computing these \commentMR{dispersion measures} for each pixel they \commentMR{are} aggregated over each segment using different schemes like mean/variance over boundary/inner pixels and relative quantities between these. This results in a total of 75 metrics for each segment which are used as input for the meta segmentation network to predict the \commentMR{segment-wise IoU} \commentMR{(for further details we refer to \cite{metaseg} and \cite{RottmannS19})}. This approach has the advantage that predicted segments are rated as a whole whereas other methods that predict the confidence pixel wise, such as \cref{eq:entropy,eq:probdiff,eq:varrat}) \commentMR{(hence providing uncertainty heat maps),} typically have a concentration of low confidence at the boundary of objects.


\commentMR{Recalling the assumption that unknown concepts are coming with suspicious segmentations, we detect predicted segments with low estimated IoU values below a chosen threshold.}
\commentMR{Subsequently, each detected segment is framed by a bounding box, i.e., the rectangular box containing all pixels of the predicted segment with \emph{minimal} width and height. The corresponding crops of the original image are then subject to further processing and retrieval analysis.}

\section{Retrieval}\label{sec:retrieval}

\commentMR{After detecting image crops corresponding to segments with low estimated quality (potential unknown OOD objects) from newly collected data, further exploration of these crops can reveal weaknesses of the CNN with respect to the given domain shift. One promising approach is content-based image retrieval \cite{BabenkoSCL14} which can help finding similarities in the crops and ultimately rate the relevance of clusters with low effort.}


Image retrieval is a well known problem with numerous applications in, \eg, search engines, automatic 3D reconstruction or document analysis. Typically retrieval starts with a query image which acts as an anchor to sort the available data points. Ranking is almost always done by calculating a visual similarity and sorting all available samples according to the similarity value. Thus there are two decisions to make: First one needs to extract visual features and then apply a distance function.

For extracting visual features we evaluate a \emph{VGG16} network \cite{vgg}, different sizes of a \emph{ResNet} \cite{resnet}, \emph{WideResNet} \cite{wideresnet} and \emph{DenseNet} \cite{densenet} all pretrained on \emph{ImageNet} \cite{imagenet} (implementation and pretrained weights are taken from the PyTorch \cite{pytorch} library). \commentMR{For all these networks we remove the final fully connected layers, i.e., we only use their backbones for feature generation.}
Unlike the other evaluated architectures, the original \emph{VGG} network \commentMR{does not include} global average pooling before the fully connected layers. To be able to extract features \commentMR{of a fixed dimensionality, independently of the size of the input image,} we also perform global average pooling \commentMR{to the \emph{VGG} backbone}.
\commentPO{As all evaluated network architectures have a limit on the minimum input size we only detect segments with a predefined minimum bounding box height and width.}
Another necessity \commentPO{for this choice} is that the visual information in a small window of the image would be very low which is not beneficial for grouping objects based on visual similarity.

\commentMR{After computing feature vectors of image crops -- which we term \emph{embeddings} -- we desire to explore their similarities. The most common and intuitive distance function is probably the \emph{euclidean distance} (\cref{eq:l2})}. Another frequently used metric is the \emph{cosine similarity} (\cref{eq:cos}) which can be beneficial for high dimensional data \cite{PerronninLSP10} like the feature vectors extracted from neural networks\commentMR{,}
\begin{align}
    L^2(x, y) &= \sqrt{\sum_{i=1}^n (x_i - y_i)^2}\label{eq:l2}\quad x,y\in\mathbb{R}^n \, ,\\
    \cos(x, y) &= \frac{\sum_{i=1}^n x_iy_i}{\sqrt{\sum_{i=1}^n x_i^2}\sqrt{\sum_{i=1}^n y_i^2}}\quad x,y\in\mathbb{R}^n\setminus\{0\} \, .\label{eq:cos}
\end{align}
In order to reduce noise in the computed embeddings and to focus on features that are relevant for the detected objects, a dimensionality reduction can improve retrieval performance significantly. To accomplish this we will evaluate in \cref{subsec:retrieval_eval} different numbers of dimensions. For a dimension lower than four we will utilize the t-distributed stochastic neighbor embedding (t-SNE) \cite{tsne} method. It is a method specifically designed for visualizing high dimensional data in low dimensional spaces while minimizing the Kullback-Leibler divergence between joint probabilities of the reduced and original space. For a more detailed explanation \commentPO{we refer to} \cite{tsne}. All other dimensionality reductions are performed using principal component analysis (PCA).

In summary, our complete OOD detection and retrieval pipeline looks as follows:
\begin{enumerate}
    \item Gather semantic segmentation predictions of newly collected samples.
    \item \commentMR{Rate all predicted segments according to their IoU estimated by MetaSeg (trained solely on the in-distribution domain).}
    \item Detect segments with estimated IoU $<0.5$ that are, however, still predicted to belong to classes that are of high interest \commentPO{\footnote{In our experiments we focus on the classes wall, fence, traffic light, traffic sign, person, rider, car, truck, bus, train, motorcycle and bicycle.}}.
    For each candidate segment, the corresponding bounding box is used to provide a crop of the original input image.
    \item Feed each crop through an embedding network pretrained on ImageNet and compute \commentMR{vectors of visual features}.
    \item (Optional) Reduce the dimensionality of the embedding space.
    \item Perform retrieval by nearest neighbor search in the resulting embedding space.
\end{enumerate}

\section{Experimental Evaluation} \label{sec:eval}

For our experimental evaluation we use a state-of-the-art DeepLabv3+ semantic segmentation model. The implementation and pretrained weights on the Cityscapes training dataset are taken from the GitHub repository of \cite{semantic_cvpr19}. The architecture uses a WideResNet38 backbone and achieves \commentMR{a} mIoU of $83.5\%$ on the Cityscapes test dataset on the standard label set and an mIoU of $92.2\%$ on the category labels. We use the Cityscapes training set as source domain and the A2D2 dataset \cite{a2d2Dataset} as target domain in which we try to find \commentPO{objects that are not well represented in the source domain.}
\commentMR{While Cityscapes only contains urban street scenes, A2D2 in addition provides scenes from highways and countryside. Therefore A2D2 is a suitable choice for exploring concepts not contained in Cityscapes.}
\commentMR{In fact, there are classes in A2D2, for example \emph{tractor} or \emph{obstacles / trash}, that are not present in Cityscapes. The class \emph{road blocks} has} a large portion in common with the Cityscapes classes \emph{fence}, \emph{wall} and \emph{guard rail} but also contains ``highway fences'' from German highways that are not present in urban environments.
This makes A2D2 highly suitable as a target domain to simulate newly collected data that has to be explored and analysed in terms of domain shift and possible new object classes that should be integrated into the next version of the model. \commentMR{As A2D2 provides quite a large number of images ($30\,000$ in total),} we randomly sample $2\,000$ images and additionally include all images \commentMR{that contain} instances of the \emph{tractor} class as this class is completely new to the semantic segmentation model. This leaves us with approximately $2\,100$ images.

All models and experiments were implemented using the PyTorch framework \cite{pytorch} and the source files can be found in our GitHub repository\footnote{\url{https://github.com/RonMcKay/OODRetrieval}}.

\subsection{Out-of-Distribution Detection}\label{subsec:metaseg_eval}
\begin{figure}[!tb]
    \centering
    \includegraphics{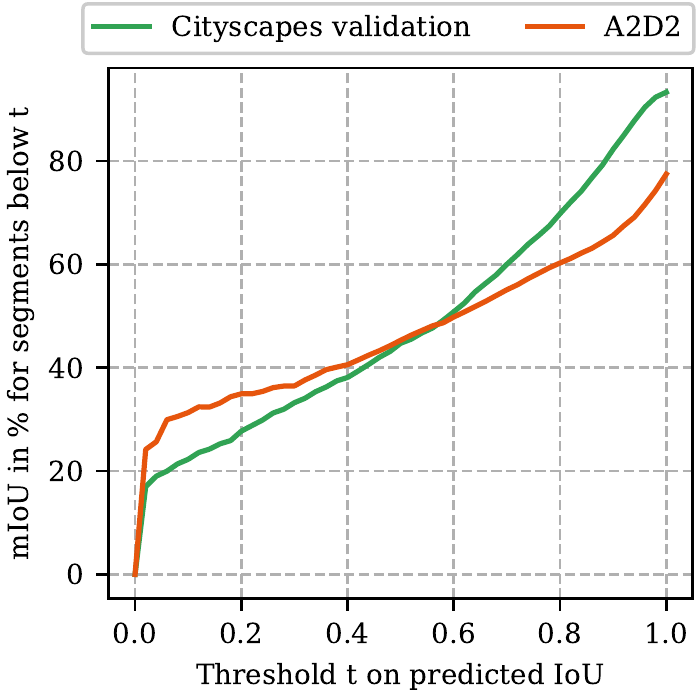}
    \caption{Intersection over union for segments that have a predicted IoU below a given threshold.}
    \label{fig:iou}
\end{figure}
\begin{figure*}[!tb]
    \centering
    \includegraphics[width=0.495\textwidth]{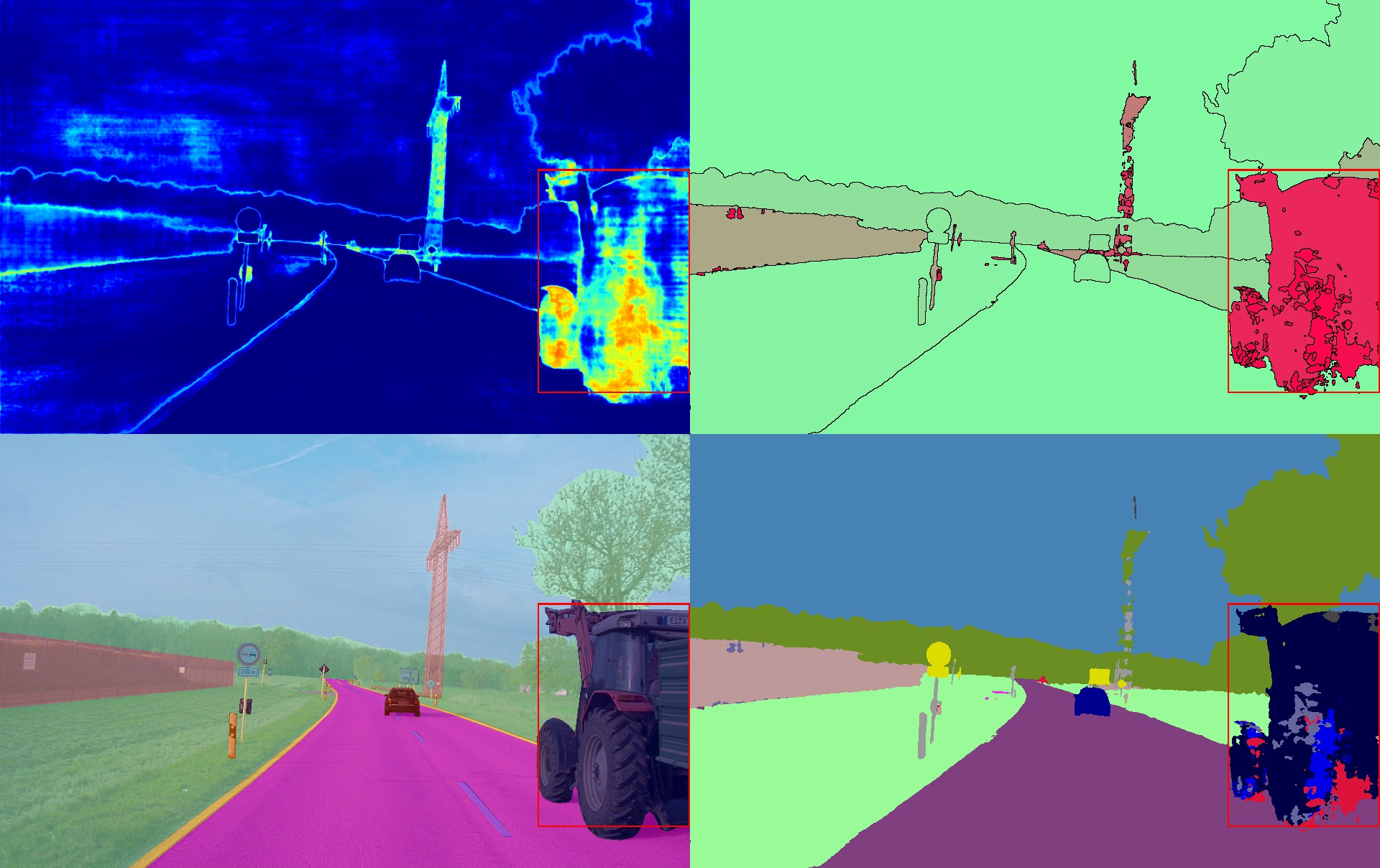}
    \includegraphics[width=0.495\textwidth]{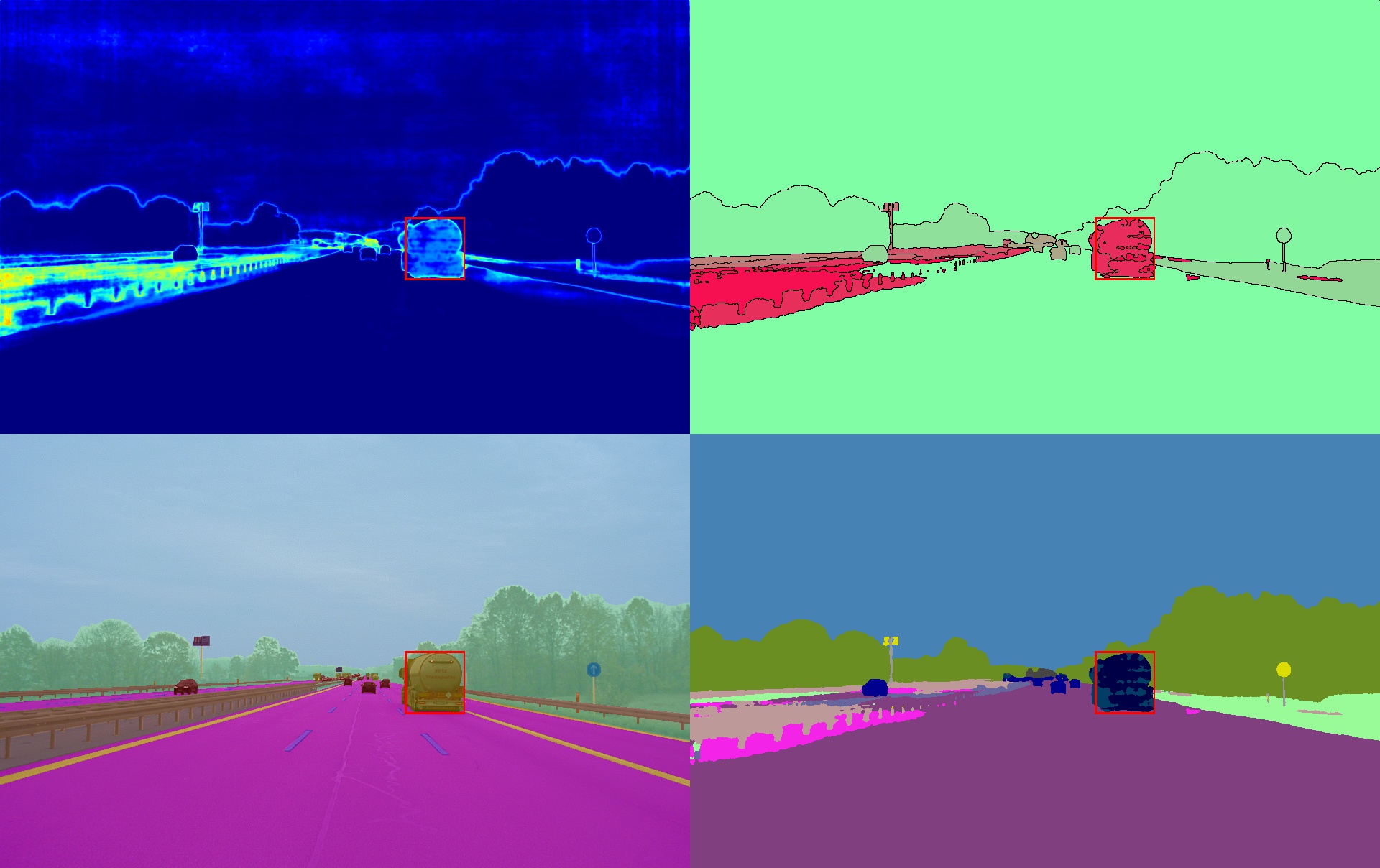}
    \caption{Two sample images from the A2D2 dataset. \commentPO{Each of them consists of four panels}. Top left: Per pixel entropy heatmap, top right: prediction of MetaSeg (green color represents high predicted IoU values, red represents low ones), bottom left: annotation over input image, bottom right: predicted semantic segmentation.}
    \label{fig:preds}
\end{figure*}
\begin{figure}[!tb]
    \centering
    \includegraphics{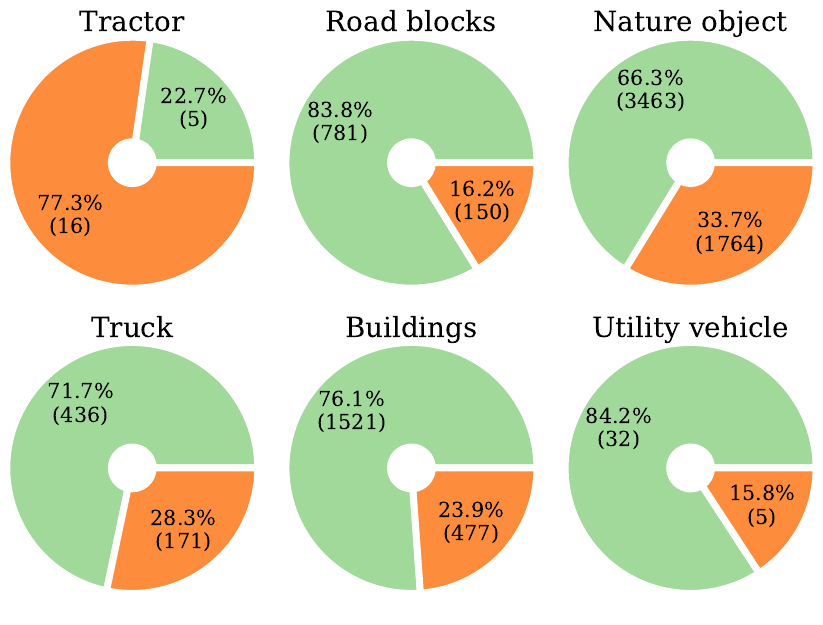}
    \caption{Number of (not) detected instances of selected classes from the A2D2 dataset. The minimum size was set to $128\times 128$ pixels, green: not detected, red: detected.}
    \label{fig:instance_counts}
\end{figure}
\begin{figure}[!tb]
    \centering
    \includegraphics{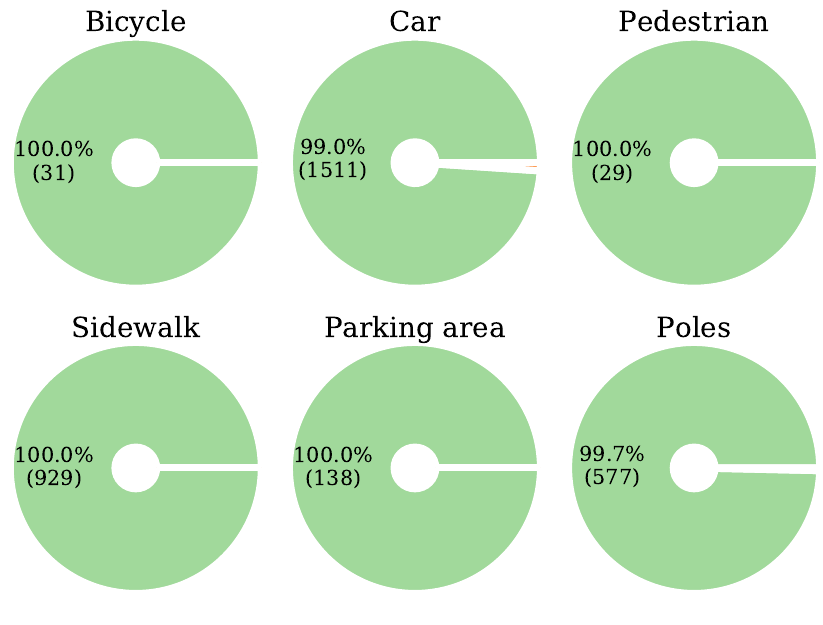}
    \caption{Number of not detected instances of selected classes from the A2D2 dataset. The minimum size was set to $128\times 128$ pixels, green: not detected, red: detected.}
    \label{fig:instance_counts_minority}
\end{figure}
So far the performance of \commentMR{MetaSeg} \cite{metaseg} has not been evaluated on datasets that are different from the domain \commentMR{MetaSeg} has been trained on. To \commentMR{demonstrate} the suitability of MetaSeg to find badly segmented objects on out of domain samples, we evaluate the DeepLabv3+ model \cite{semantic_cvpr19} on the A2D2 dataset.
\commentPO{\commentMR{First, we calculate}
the mIoU of the semantic segmentation model on the target domain to have a reference in terms of segmentation quality. \commentMR{However,} the label sets of A2D2 and Cityscapes are not compatible, as discussed earlier. 
} \commentMR{Hence, we perform}
a label mapping from the full A2D2 label set to the Cityscapes training set. To minimize mapping errors we further map the Cityscapes classes onto their coarse category ids. Excluding the \emph{void} class this leaves us with the seven categories \emph{flat}, \emph{construction}, \emph{object}, \emph{nature}, \emph{sky}, \emph{human} and \emph{vehicle}. Note that we made a small tweak \commentPO{where we} mapped the \emph{rider} class of Cityscapes to the \emph{vehicle} instead of the default \emph{human} category. \commentMR{This is motivated by the different annotation styles, i.e., a person riding a bicycle is annotated as bicycle in the A2D2 dataset which is in contrast to the annotation in Cityscapes where the person is annotated as a \emph{rider}.}
\commentMR{DeepLabv3+ achieves a remarkable test accuracy of $99.2\%$ on the Cityscapes test set with respect to the coarse categories.}
\commentPO{Evaluating the DeepLabv3+ on the A2D2 dataset with this label mapping \commentMR{still} results in a mIoU of $77.4\%$.}

In order to \commentMR{demonstrate} the effectiveness of MetaSeg we compute the mIoU under different thresholds, removing segments with an \commentPO{estimated} IoU above the specified threshold from the evaluation.
This leads to a minimal performance of around $20\%$ as \commentMR{shown} 
in \cref{fig:iou}. Although the thresholding does not work as nicely as for the Cityscapes validation set, the experiment shows that we are able to identify badly segmented regions and 
\commentMR{detect} 
them confidently by \commentMR{means of the} predicted IoU.

Two example predictions can be seen in \cref{fig:preds}. On the left the unknown \emph{tractor} object \commentMR{is} segmented very poorly which \commentMR{is} detected by MetaSeg. On the right the truck as well as the highway fence are badly segmented which \commentMR{is} detected as well.

\commentMR{Next,} we performed an experiment to test \commentMR{whether} unknown objects are consistently segmented with \commentMR{low} quality, i.e., low IoU, and also if \commentMR{MetaSeg} predicts a low IoU. This would be an indispensable property to be able to find these objects reliably. \commentMR{Under the assumption that large objects with low predicted IoU are most critical,} we collect all segments that have a predicted IoU of less than $0.5$ and a minimum size of \commentPO{$128\times 128$} pixels and count how many instances of each class are covered by at least one of these segments. We count an instance as \emph{covered} by a segment when the total number of pixels of that instance inside the segment is at least $50\%$ of the total segment size. \commentMR{For the evaluation we also only consider ground truth instances that have} a minimum size of \commentPO{$128\times 128$} pixels, \commentMR{for a given class their number represents the amount of instances}. 
Note that, in this scenario we do not use the label mapping as we want to explore the behavior of our approach with respect to all A2D2 classes. \commentMR{\Cref{fig:instance_counts} shows that MetaSeg detects $77.3\%$ of the instances belonging to the \emph{tractor} class (which is a class not present in the source domain).} \commentPO{Reviewing the \commentMR{remaining} $22.7\%$ of \emph{tractor} instances that were not detected} they are mostly in scenarios where the tractor was obscured to a large degree by other vehicles or in situations where the tractor was not on the road but on a nearby farming field. When \commentPO{analyzing} the $16.2\%$ of \emph{road blocks} instances, \commentMR{we observe that they are} almost exclusively segments of ``highway fences''. The classes \emph{nature object} as well as \emph{truck} have also a relatively high share of detected segments. \commentPO{This is} due to the fact that large trucks and forests / grasslands that cover a big portion of the image are rather rare in urban environments. The \emph{building} cluster consists to a large degree of tunnels and bridges that span \commentPO{the street}.
\commentPO{\Cref{fig:instance_counts_minority} shows classes that are common in the Cityscapes dataset.} \commentMR{The fact that almost no segments belonging to these classes are detected (except for a few cars, sidewalks and poles) further demonstrates the performance of MetaSeg on this out-of-distribution detection task.}
\commentMR{Performing this same evaluation on the Cityscapes test set leads to an average of $0.06$ detected segments per image whereas we detect $0.82$ segments per image in the A2D2 dataset. This shows that we are consistently detecting out of distribution objects.}

\subsection{Retrieval Task}\label{subsec:retrieval_eval}
\begin{figure*}[!tb]
    \centering
    \includegraphics{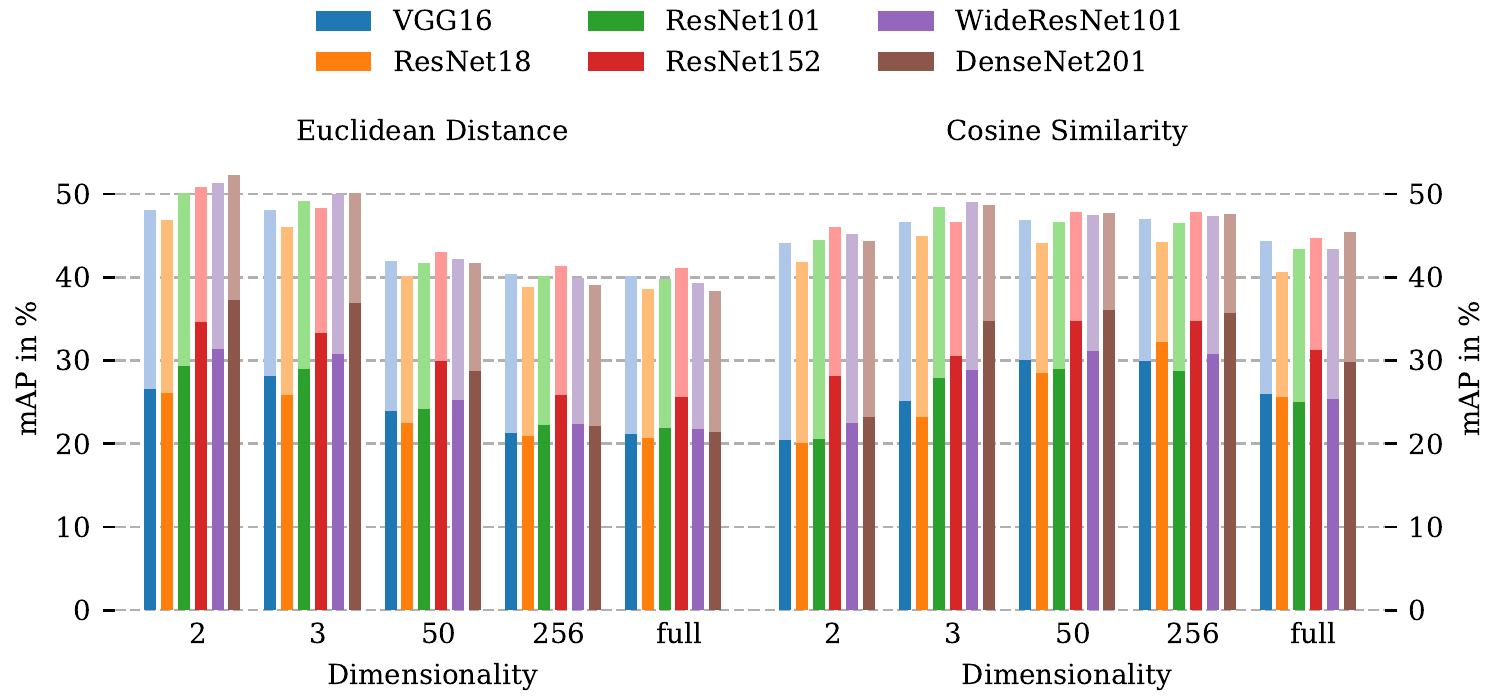}
    \caption{Retrieval results for different feature extractors, distance metrics and feature dimensionalities. For dimension two and three PCA down to 50 dimensions followed by t-SNE \cite{tsne} has been used for dimensionality reduction. For the dimensions 50 and 256 PCA has been performed. A dimension of \emph{full} means no dimensionality reduction. Transparent bars correspond to mean over all queries not taking their class into account thus being weighted by frequency. Non transparent bars correspond to the mAP when averaging over all queries \commentMR{first class-wise} and then over all classes.}
    \label{fig:maps}
\end{figure*}
\begin{figure}[!tb]
    \centering
    \includegraphics{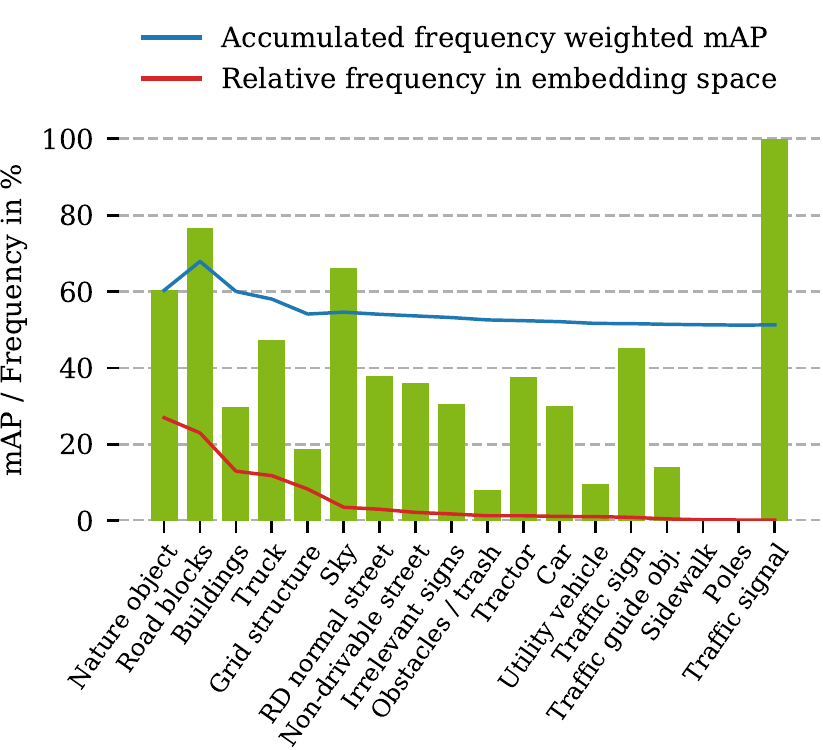}
    \caption{\commentPO{Class-wise retrieval results for a \emph{DenseNet201} feature extractor using PCA to reduce to 50 dimensions followed by t-SNE \cite{tsne} to reduce the embedding dimensionality to two dimensions. For measuring similarity the Euclidean distance has been used. The classes have been sorted from left to right in descending order according to their frequency in the embedding space. Results are measured in mAP \commentMR{percentage}. 
    }}
    \label{fig:class_maps}
\end{figure}
\commentPO{In this section we present a qualitative and quantitative evaluation of the retrieval task described in \cref{sec:retrieval}.
\commentMR{\Cref{fig:embedding_space_density} depicts an embedding space of features computed by a \emph{ResNet152}.} Dimensionality reduction has been performed using principal component analysis down to $50$ dimensions followed by \emph{t-SNE} \cite{tsne} (the original feature space had a dimensionality of 2048).}
\commentMR{Therefore, each predicted segment that has been detected by MetaSeg is mapped to a data point / sample in $\mathbb{R}^2$.}
Qualitatively the space seems to be well separated into different clusters of objects. \commentMR{Note that this is achieved without using any ground truth of A2D2}.
\commentMR{In order to visualize that the embedded features of detected segments are clustered according to their semantics, we assign the A2D2 color code to the embedded samples. Herein, the class assigned to the predicted segment is the class of the ground truth segment that has maximal overlap with the predicted segment.}

\commentPO{The classes \commentMR{that} \commentMR{contribute} the majority of \commentMR{data points in} the embedding space are \emph{nature object}, \emph{road blocks}, \emph{buildings} and \emph{truck}. All of them are well separated into individual clusters. The \emph{tractor} class, although having a relative share of only $1.3\%$ does also form a cluster close to trucks, which are semantically related.}
Embeddings with the ground truth class \emph{sky} seem to emerge from situations with \commentMR{borderline} 
weather conditions like rain or direct sunlight. \commentMR{In both cases the neural network tends to make 
false predictions within sky regions. In addition,} one can notice that in highway scenes the sky covers a much larger area than in urban scenes. 
\commentMR{However, due to the training data consisting of urban scenes, the segmentation network is potentially biased towards predicting buildings in case the sky looks unusual.}


For measuring retrieval performance quantitatively we use the \emph{mean average precision} (mAP) which is defined as
\begin{align}
    \text{mAP} &= \frac{1}{Q} \sum_{q=1}^Q \text{AP}_q \, ,\quad\text{\commentPO{where}}\\
    \text{AP}_q &= \frac{\sum_{i=1}^np(i)\cdot r(i)}{t} \, ,
\end{align}
with $p(i)$ being the precision when cutting off the retrieval list at position $i$, $r(i)$ an indicator function equaling $1$ if element $i$ is relevant with respect to the query and $0$ otherwise, $n$ is the total number of data points and $Q$ the total number of queries.
It holds that $\text{mAP}\in(0,1]$, being $1$ if all relevant objects are at the top of the retrieval list for each query and minimal if all relevant objects are at the end of each retrieval list.

In \cref{fig:maps} the quantitative retrieval results \commentPO{with respect to different feature extractors, embedding space dimensionalities and distance metrics} are summarized in terms of mAP. The Euclidean distance benefits from the t-SNE embedding into a lower dimensional space. This is in contrast to the cosine similarity that performs worse in low dimensional spaces. \commentMR{However, the results for cosine similarity show a small performance increase when going from two to three dimensions. Over the range of dimensions, the cosine similarity results appear to be more stable.} 
Regarding the different feature extractors, deeper networks with more filters extract more meaningful features which are better in retrieving visually similar objects. The overall best performing network is the \emph{DenseNet201} with $52.2\%$ mAP followed by the \emph{WideResNet101} with $51.3\%$ mAP. The large gaps between global and class wise mAP is due to a few \commentMR{under represented} classes like, \eg, \emph{poles} or \emph{sidewalk}. In our experiments they have only a few samples and a high visual variability due to many distracting background objects. This is why these classes get a mAP in the range of $0.16 - 0.38\%$ and reduce the average over classes. \commentPO{\Cref{fig:class_maps} depicts the mAP values for the best performing setup (\emph{DenseNet201} with PCA/t-SNE reduced to two dimensions) split up into the different classes that are present in the embedding space. The results are sorted from left to right in descending order according to their frequency in the embedding space.}
\commentMR{In practice, the frequency of an unknown object is very likely to be an important indicator. Therefore, it should be estimated before making a decision whether to acquire new data for training. Our retrieval results in terms of mAP show that retrieval is useful for such an estimation and for data selection in general.}

\begin{figure}[tb]
    \centering
    \includegraphics[width=\columnwidth,keepaspectratio]{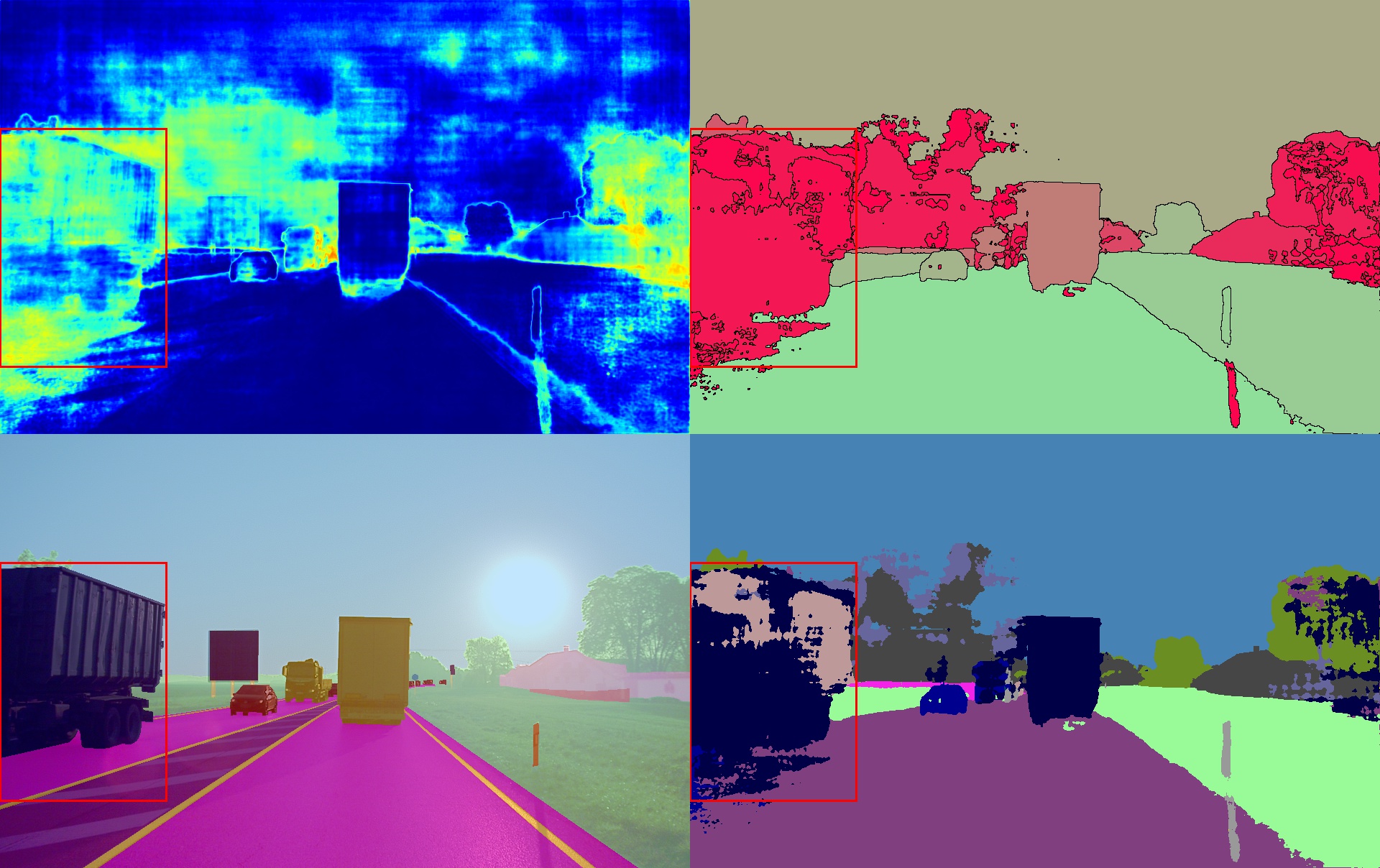}
    \caption{Sample prediction from the A2D2 dataset. \commentPO{Panels} are the same as in \cref{fig:preds}. The trailer on the very left hand side is labeled as \emph{tractor}, presumably based on previous frames. \commentMR{This can hardly be established without utilizing temporal information or expert knowledge.}
    The \commentMR{overall} badly predicted segmentation \commentMR{is likely caused by recording} against direct sunlight.}
    \label{fig:trailer}
\end{figure}

Intuitively the qualitative results in \cref{fig:embedding_space_density} seem to be better than the achieved mAP results from \cref{fig:maps}. The reasoning behind this is as follows. When looking at the label set of A2D2, which is where we extract the ground truth information for the retrieval task, the visual variability of some classes is too high to perform retrieval on them based on visual similarity. The class \emph{utility vehicle}, \eg, contains not only trams but also excavators and other construction machines which have a rather low visual similarity. Another example is the \emph{tractor} class. \commentPO{\Cref{fig:trailer} shows a trailer on the \commentMR{very left hand} side.} Without any \commentPO{further} context, classifying this trailer to belonging to a tractor is challenging. Previous frames however \commentMR{reveal} that the trailer is mounted to a tractor, \commentMR{therefore it is presumably labeled as tractor.}
\commentPO{In order to compute visual features that are correlated with the other \emph{tractor} class instances, \commentMR{sequential models could be considered for exploiting correlations in consecutive frames.} The described issues only represent a few of the challenges that we face when extracting ground truth information from pixel label annotation for evaluation of the retrieval task. \commentMR{In general the retrieval evaluation suffers from the coarse semantic classes and inconsistencies among the datasets. Datasets that provide a label set with more fine-grained object classes might increase the quantitative retrieval results significantly.}}
Nonetheless, the qualitative and quantitative results show that the embedding space is suitable for exploring newly collected data and that the proposed pipeline can be used to accelerate feedback from the deployment phase to an update of the training dataset. \commentPO{Note that the proposed pipeline \commentMR{is of generic nature, in a sense that the user has the freedom to choose any OOD detection method as well as any semantic segmentation model.}}

\section{Conclusion and future work}

In this work we have demonstrated and validated how to use \commentMR{prediction quality estimation} methods, such as MetaSeg, and image retrieval to explore newly collected data that might be affected by domain shift. We are able to detect object classes that are unknown to the semantic segmentation network due to missing samples in the training set. Data exploration can be guided by image retrieval on visual features that are gathered by common deep learning architectures which are trained on the task of image classification. \commentMR{However, dataset selection for evaluating this kind of methods leaves room for further improvement. Also benchmarking in the fields of OOD detection and uncertainty on OOD samples in semantic segmentation remains tedious due to the lack of appropriate datasets. We believe that these subjects deserve to be further addressed in the future.}

In terms of future work, we plan to explore possibilities to utilize the detected segments. Methods like \emph{active learning} or \emph{semi supervised learning} can be used to reduce annotation cost for new object classes and still incorporate them into a new training set. The knowledge gained from a human in the loop in an active learning setting could also be used to automatically retrain the embedding network. This way the visual features would be more meaningful in the context of the current environment and thus increase retrieval performance. \commentPO{Another possible research direction is the utilization of temporal correlation between adjacent frames or segmentation networks that are trained to be uncertain on unlabeled classes like, \eg, the void classes of Cityscapes.}


\section*{Acknowledgment}

This work is in part funded by the German Federal Ministry for Economic Affairs and Energy (BMWi) through the grant 19A19013Q, project AI-DeltaLearning. Furthermore we thank Hanno Gottschalk for useful advice and discussion.


{\small
\bibliographystyle{ieee_fullname}
\bibliography{main}
}

\end{document}